\documentclass[letterpaper, 10 pt, conference]{ieeeconf}  % Comment this line out if you need a4paper

\IEEEoverridecommandlockouts                              % This command is only needed if 
\usepackage{amsmath} % assumes amsmath package installed
\usepackage{graphicx}
\usepackage[algorithm]{algorithm}
\usepackage{algorithmicx}
\usepackage{algpseudocode}

\title{\LARGE \bf
Executor-aware Candidate Selection via a Feasibility Certificate
}
 
\author{Sooin Choi$^{1}$, Soonwoong Hwang$^{2}$ and Wansoo Kim$^{2*}$% <-this  %stops a space
\thanks{$^{1}$The author is with Department of Robotics, Hanyang University, Seoul, Republic of Korea
      {\tt\small sooince@hanyang.ac.kr}}%
\thanks{$^{2}$The author is with Department of Robotics, Hanyang University ERICA, Ansan, Republic of Korea
       {\tt\small hswfile@hanyang.ac.kr},
    {\tt\small wansookim@hanyang.ac.kr}        }
 \thanks{$^{*}$Corresponding author: Wansoo Kim.}%
}

\begin{document}
 
\maketitle
\thispagestyle{empty}
\pagestyle{empty}

\begin{abstract}
 
Modular robotic systems often separate motion planning from a downstream executor that enforces state-dependent hard constraints. A candidate that is geometrically valid may therefore be incompatible with the executor's available command set. We present a certificate-based candidate-selection framework that constructs a command witness from the executor hard set at predicted rollout states and verifies it against the original constraints, without changing candidate generation, ranking, or the executor.
 
Across 5,085 geometry-valid numerical evaluations on two robot models, 795 admitted no executor-feasible command. The certificate is sufficient but conservative: none of the 795 was certified, while 7.09\% of reference-feasible cases remained uncertified. In controlled FR3 and fixed-base RB-Y1 simulations, certificate admission frequently changed candidate selection, and a post-hoc exact linear-programming (LP) admission baseline revealed platform-dependent conservatism. Relative to geometry-based selection, certificate admission was associated with lower planner-command coverage and higher nominal tracking error, without a consistent advantage in reached-state interaction reserve. A planner-generated MoveIt/OMPL study further evaluates the same admission rule on externally generated candidate pools. 
%Under the tested single-thread implementation, median state-level check times were 158 and 140~$\mu$s for FR3 and RB-Y1, compared with 669 and 627~$\mu$s for a zero-objective feasibility LP over the same hard set.
\end{abstract}
%%%%%%%%%%%%%%%%%%%%%%%%%%%%%%%%%%%%%%%%%%%%%%%%%%%%%%%%%%%%%%%%%%%%%%%%%%%%%%%%

\section{INTRODUCTION}
\label{sec:introduction}
 
As robots increasingly operate in shared environments and physically interact with humans, satisfying safety and interaction constraints while maintaining task performance has become a central challenge\cite{Haddadin2012Safety}. Hierarchical control\cite{Sentis2005WholeBody,Escande2014HQP} is widely adopted for this purpose, allowing lower-priority tasks to be performed subject to higher-priority safety requirements. In such architectures, safety and environmental constraints can be assigned higher priority, while task-oriented objectives are pursued at lower priority.
 
In modular robotic systems, hierarchical controllers can operate as executors downstream of a separate motion planner\cite{Herbert2017FaSTrack}. In this architecture, the planner generates future motion candidates or references, while the hierarchical executor computes the final command based on the current state and active hard constraints. This separation can create a mismatch between the validity criteria used by the planner and the command feasibility imposed by the executor. In particular, a candidate that is valid under the planner's criteria may not admit a command satisfying the executor's active constraints. In our numerical cohort, most such incompatibilities violate a single hard row that geometric validity does not check; under an active interaction constraint, we also observe cases that arise only from the coupling of rows that are individually satisfiable over the command box (Sec.~\ref{subsec:geometric_validity_and_executor_feasibility}).
 
\begin{figure}[t]
    \centering
    \includegraphics[width=\columnwidth]{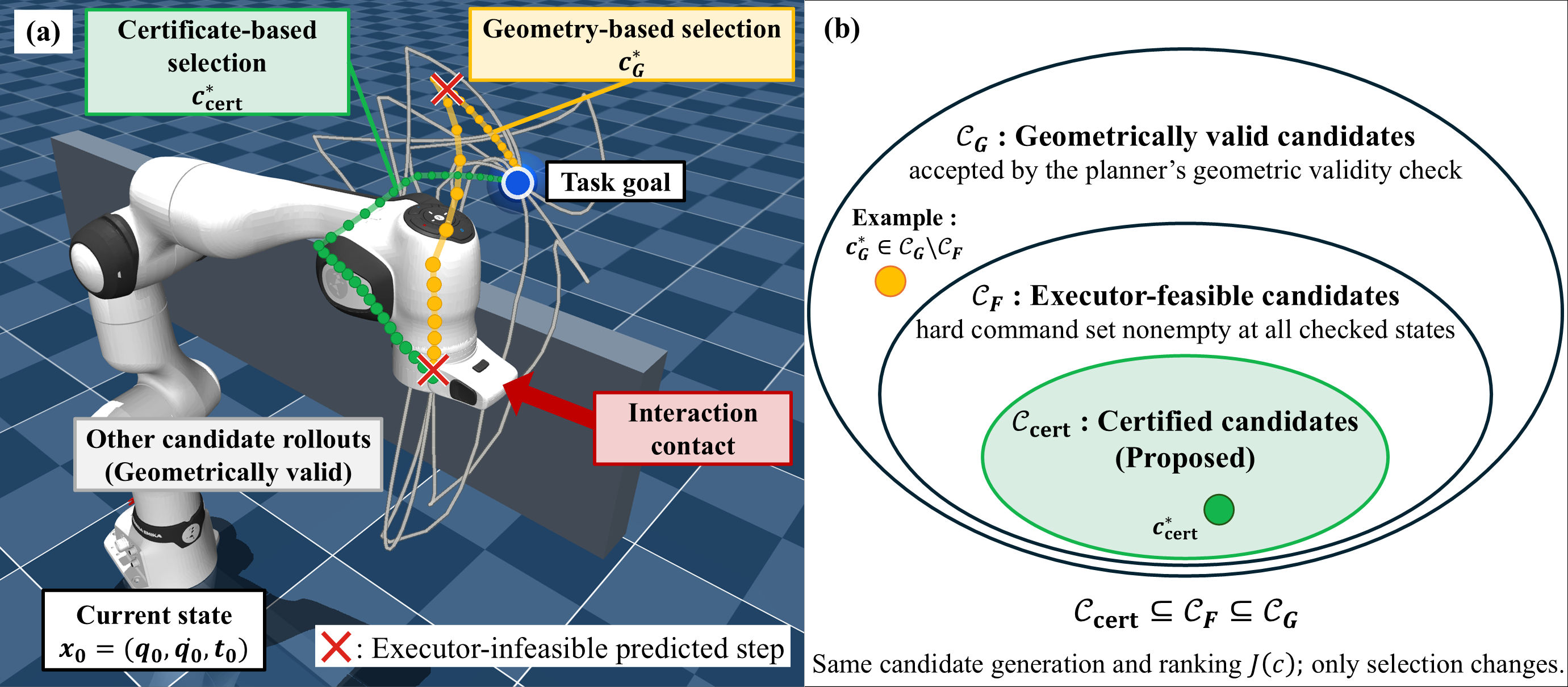}
    \caption{Concept of executor-aware candidate selection. (a) From the same current state $x_0$, the planner generates geometrically valid candidate rollouts toward a common task goal. In the illustrated example, the geometry-based selection $c_G^*$ contains an executor-infeasible predicted state, whereas certificate-based admission selects $c_{\mathrm{cert}}^*$. Candidate rollouts are visualized by their end-effector paths for clarity; certification is evaluated using the full predicted robot state and executor hard constraints. (b) Under the zero-tolerance idealization, $\mathcal C_{\mathrm{cert},0}\subseteq\mathcal C_F\subseteq\mathcal C_G$. In implementation, a constructed witness is accepted after validation against the original hard set within the residual tolerance $\epsilon_{\mathrm{cert}}$; candidate generation and ranking remain unchanged.}
\label{fig:overview} \vspace{-1.2em}
\end{figure}
 
Among various planning approaches, classical geometric planners such as RRT\cite{Kuffner2000RRT} and PRM\cite{Kavraki1996PRM} generate candidate trajectories based on collision avoidance and configuration-level constraints, such as joint-position limits. Previous studies have addressed feasibility at several levels to solve this problem. Kinodynamic planners extend geometric planning by incorporating system dynamics and control-input constraints into the planning process, thereby generating dynamically realizable trajectories\cite{Webb2013KinodynamicRRTStar}. However, the resulting feasibility is defined by the planner's own dynamics and input models and therefore need not coincide with the command feasibility imposed by a separate hierarchical executor with additional state-dependent hard constraints. Controller-aware approaches such as FaSTrack\cite{Herbert2017FaSTrack} and Funnel Libraries\cite{Majumdar2017FunnelLibraries} select trajectories based on tracking-error bounds or regions that can be maintained by a feedback controller. These methods explicitly improve planner-executor compatibility, but mainly target tracking robustness rather than directly evaluating the feasible command set defined by a hierarchical executor.
 
In human-robot interaction, interaction-aware approaches \cite{Haddadin2012Safety,Flacco2012DepthSpace} add state-dependent requirements from human proximity and contact, which can restrict the executor's command set when active even if the planned motion remains geometrically valid.
 
Other approaches handle feasibility by changing where the planning and control problem is solved. Model predictive control (MPC)\cite{Rawlings2017MPC} integrates dynamics, state constraints, and input constraints within a single optimization problem, thereby avoiding a separate planner--executor feasibility interface. Within hierarchical controllers, feasibility can be handled by modifying the executor-side optimization, for example, task or constraint relaxation\cite{DiLillo2019SetBasedMultiTaskIK}, control-barrier-function-based
hierarchical quadratic programming (CBF-HQP)\cite{Xie2025CBFHQP}, or online task-priority and redundancy modulation\cite{Chen2024DynamicMultiTaskPriority,Kanoun2009ConstraintPrioritization}.
Optimization-based safety filters such as CBF-QP controllers encode safety
requirements as constraints on admissible controls while optimizing a nominal performance objective\cite{Ames2017CBFQP}; feasibility under simultaneous control and safety constraints has also been studied explicitly\cite{Xiao2022CBFFeasibility}. At a broader dynamical level, viability theory addresses whether admissible future evolutions can remain within a prescribed constraint set\cite{Aubin1991Viability}. In contrast, the present certificate is a state-wise admission test for the hard-command set of a separately deployed downstream executor; it neither modifies that executor nor establishes forward invariance or viability. The remaining challenge considered here lies at the planner--executor interface: determining whether a planner-generated candidate is compatible with the active hard-command set of a fixed executor.
 
Accordingly, we propose a certificate-based candidate-selection framework that evaluates executor-level command feasibility at predicted rollout states while preserving a fixed hierarchical quadratic programming (HQP)\cite{Escande2014HQP} executor (Fig.~\ref{fig:overview}). The certificate itself uses only the executor hard-command set and does not depend on the soft HQP task hierarchy; the HQP considered here is the experimental executor instance. The geometry-based selector ranks all geometrically valid candidates in $\mathcal C_G$, whereas certificate-based admission restricts selection to $\mathcal C_\mathrm{cert}$; under the zero-tolerance idealization, the corresponding set $\mathcal C_{\mathrm{cert},0}$ satisfies $\mathcal C_{\mathrm{cert},0} \subseteq\mathcal C_F$.

\section{METHODOLOGY}
\label{sec:methodology}
 
\subsection{Executor Feasibility}
\label{subsec:excutor_feasibility}
 
\begin{figure}[t]
    \centering
    \includegraphics[width=\columnwidth]{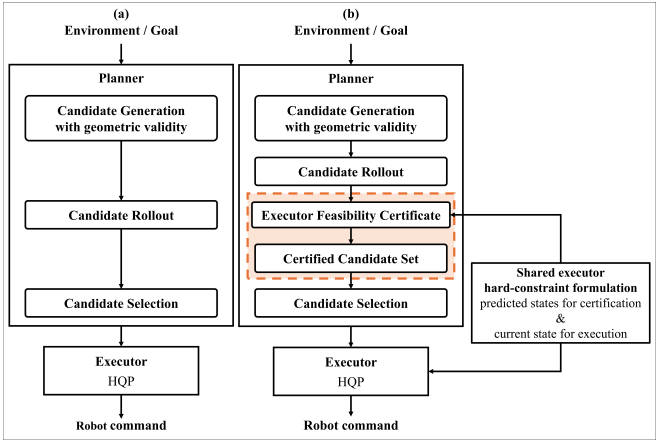}
    \caption{Geometry-based and executor-feasibility-aware candidate selection. The proposed framework evaluates the executor hard-constraint formulation at predicted rollout states before selection, while the fixed HQP executor applies the same formulation at the current state.} \vspace{-1.2em}
    \label{fig:architecture}
\end{figure}
 
We consider a hierarchical executor that generates a joint velocity command $u$ from the state $x=(q,\dot q,t)$ (Fig.~\ref{fig:architecture}), first restricted by physical joint command bounds:
\begin{equation}
\begin{aligned}
\ell_i=\max\{&
-v_i^{\max},\;
\dot q_i-a_i^{\max}\Delta t,\\
&-\frac{q_i-q_i^-}{\Delta t},\;
-\sqrt{2a_i^{\mathrm{br}}[q_i-q_i^-]_+}
\},
\end{aligned}
\label{eq:lower_velocity_bound}
\end{equation}
\begin{equation}
\begin{aligned}
h_i=\min\{&
v_i^{\max},\;
\dot q_i+a_i^{\max}\Delta t,\\
&\frac{q_i^+-q_i}{\Delta t},\;
\sqrt{2a_i^{\mathrm{br}}[q_i^+-q_i]_+}
\}.
\end{aligned}
\label{eq:upper_velocity_bound}
\end{equation}
 
These bounds act directly on the commanded joint velocity and combine velocity, acceleration, one-step position, and braking-distance limits, where $q_i^{\pm}$ are the joint-position bounds, $v_i^{\max}$ and $a_i^{\max}$ are the velocity and acceleration limits, and $a_i^{\mathrm{br}}$ is the available braking acceleration\cite{Flacco2012HardJointConstraints,DelPrete2018JointBounds}; the square-root terms follow from $v^2\le2ad$ and reduce the admissible velocity as a joint approaches its boundary. Together they define the physical command box $\ell\le u\le h$, which the executor intersects with all active hard rows:
\begin{equation}
    \begin{aligned}
        \mathcal F_{\mathrm{exec}}(x)=\{u|\ell\le u\le h, A_{\mathrm{hard}}&u\ge b_{\mathrm{hard}},\\&E_{\mathrm{hard}}u = e_{\mathrm{hard}}\}
    \end{aligned}.
    \label{eq:executor_feasible_set}
\end{equation}
 
Here $A_{\mathrm{hard}}u\ge b_{\mathrm{hard}}$ collects the active hard inequalities imposed by the executor, such as workspace, obstacle, interaction, or robot-specific constraints, and $E_{\mathrm{hard}}u=e_{\mathrm{hard}}$ collects hard equalities when present; the state dependence of the command bounds and hard-constraint terms is omitted when clear from context. Task-tracking and posture objectives do not enter $\mathcal F_{\mathrm{exec}}(x)$ and only determine which command is preferred after feasibility has been established. The certificate therefore depends only on this hard-command set; the HQP used in our experiments is one executor instance whose soft task hierarchy plays no role in admission. Executor feasibility refers to local nonemptiness, $\mathcal F_{\mathrm{exec}}(x)\neq\emptyset$, rather than to full-trajectory feasibility.
 
Directly testing this nonemptiness requires solving a feasibility problem. Invoking the deployed executor at a predicted state does not answer it either: when the original hard set is empty, the executor falls back to a minimum-violation relaxation and can still return a command, so the returned command alone is not a feasibility label. A returned command is not a feasibility label in the converse direction either, since a solver failure on a hard-feasible set follows a different path and also yields a command. The optimal violation of that relaxation would provide one, but obtaining it requires a linear program of essentially the same constraint scale as the zero-objective feasibility LP reported in Sec.~\ref{subsec:computational_cost}. We therefore use the constructive certificate for repeated candidate-rollout checks.
 
\subsection{Feasibility Certificate}
\label{subsec:feasibility_certificate}
 
\begin{algorithm}[!t]
\caption{Constructive Feasibility Certificate}
\label{alg:feasibility_certificate}
\begingroup
\small
\begin{algorithmic}[1]
\Require State $x$, bounds $\ell,h$, hard constraints,
levels $\{\mathcal L_k\}_{k=1}^{K}$, tolerances
$\epsilon_{\mathrm{act}},\epsilon_{\mathrm{con}},\epsilon_{\mathrm{cert}}$
\Ensure $B_{\epsilon}(x)$ and $m_{\mathrm{cert}}$ when defined
 
\If{$\ell_i>h_i$ for any $i$}
    \State \Return $B_{\epsilon}(x)=0$
    \Comment{empty command box}
\EndIf
 
\State Initialize $\bar u,s$; center hard rows at $\bar u$ and pair equality rows
\State $d_{<1}\gets0$, $G_{<1}\gets\emptyset$
 
\For{$k=1,\ldots,K$}
    \State Compute $P_{<k}$ using \eqref{eq:nullspace_projector}
    \State $\mathcal T\gets\emptyset$, $d_k\gets0$, $j\gets0$
 
    \While{a row in $\mathcal L_k$ is violated by more than
           $\epsilon_{\mathrm{act}}$ at $d_{<k}+d_k$
           \textbf{and} $j<j_{\max}$}
        \State Add newly violated rows to $\mathcal T$; $j\gets j+1$
        \State Form $(G_{\mathcal T},g_{\mathcal T})$ and recompute $d_k$
               using \eqref{eq:level_correction}
        \If{$\mathcal T$ or $G_{<k}$ not satisfied within
            $\epsilon_{\mathrm{con}}$}
            \State \Return $B_{\epsilon}(x)=0$
            \Comment{construction failure}
        \EndIf
    \EndWhile
 
    \State $d_{<k+1}\gets d_{<k}+d_k$
    \State $G_{<k+1}\gets\operatorname{stack}(G_{<k},G_{\mathcal T})$
           \Comment{unchanged if $\mathcal T=\emptyset$}
\EndFor
 
\State $u_{\mathrm{cert}}\gets\bar u+d_{<K+1}$
\State Evaluate $m_{\mathrm{cert}}$ by \eqref{eq:certificate_value}
       and $\mathcal R(u_{\mathrm{cert}};x)$ by \eqref{eq:residual}
\State \Return $B_{\epsilon}(x)$ by \eqref{eq:decision},
       with $m_{\mathrm{cert}}$
\end{algorithmic}
\endgroup
\end{algorithm}
 
The certificate (Algorithm~\ref{alg:feasibility_certificate}) constructs a particular command from the executor hard rows and verifies it against the original hard set. If $\ell_i>h_i$ for any joint $i$, the physical command box is empty and $\mathcal F_{\mathrm{exec}}=\emptyset$. Otherwise, with center $\bar u=(\ell+h)/2$ and half-width $s=(h-\ell)/2$, the construction is expressed in the displacement coordinate $d=u-\bar u$. Each hard inequality row and each paired equality row is normalized to unit Euclidean norm in $d$, and rows with negligible norm are omitted, whereas final verification uses the original, unnormalized rows. Rows are processed in a fixed sequence of certificate levels; the ordering affects which feasible states are certified but does not redefine $\mathcal F_{\mathrm{exec}}(x)$ or introduce priorities among its constraints.
 
At certificate level $k$, let $G_{<k}$ denote the construction rows activated by the preceding levels, with null-space projector
\begin{equation}
    P_{<k}=I-G_{<k}^\dagger G_{<k},
    \label{eq:nullspace_projector}
\end{equation}
where $(\cdot)^\dagger$ is the Moore--Penrose pseudoinverse. With the displacement $d_{<k}$ accumulated through the preceding levels and the target rows $(G_\mathcal T,g_\mathcal T)$ activated so far within level $k$, the level correction is
\begin{equation}
    d_k=P_{<k}(G_\mathcal{T}P_{<k})^\dagger (g_\mathcal{T} -G_\mathcal{T}d_{<k}).
    \label{eq:level_correction}
\end{equation}
 
The projector $P_{<k}$ is fixed within a level; when additional rows become active, $d_k$ is recomputed from the enlarged target set rather than accumulated incrementally. This progressive null-space activation is related to saturation-in-the-null-space methods under hard joint constraints\cite{Flacco2012HardJointConstraints}, which generate the command executed at the current state; here the same machinery constructs a witness at predicted states that is verified and then discarded. Three tolerances play distinct roles: row activation ($\epsilon_{\mathrm{act}}=10^{-9}$), satisfaction of activated and previously processed rows ($\epsilon_{\mathrm{con}}=10^{-7}$), and the final residual on the original hard set ($\epsilon_{\mathrm{cert}}=10^{-9}$). Because the activated set only grows within a level, activation terminates after at most $|\mathcal L_k|$ iterations unless construction failure is returned earlier; the implementation additionally caps the inner loop at $j_{\max}=12$. After all levels, the final displacement gives $u_{\mathrm{cert}}=\bar u+d_{\mathrm{cert}}$ and the certificate value
\begin{equation}
    m_\mathrm{cert}=\min_i(s_i-|d_{\mathrm{cert},i}|).
    \label{eq:certificate_value}
\end{equation}
 
A nonnegative value, $m_{\mathrm{cert}}\ge0\Longleftrightarrow\ell\le u_{\mathrm{cert}}\le h$, places the constructed command inside the physical command box but does not verify the remaining hard rows, which are checked through the residual
\begin{equation}
    \begin{aligned}
        &\mathcal R (u;x)=\max\{0,\max_i(\ell_i-u_i), \max_i(u_i-h_i),\\&\max_j(b_{\mathrm{hard},j}-[A_\mathrm{hard}u]_j), \max _k|[E_\mathrm{hard}u-e_\mathrm{hard}]_k|\},
    \end{aligned}
    \label{eq:residual}
\end{equation}
where the term of an absent constraint group is omitted. The implemented state-level decision is
\begin{equation}
    B_\epsilon(x) = \begin{cases}
        1,&\ m_\mathrm{cert}\ge0,\ \mathcal R(u_\mathrm{cert};x)\le \epsilon_\mathrm{cert}\\
        0,&\mathrm{otherwise.}
    \end{cases}
    \label{eq:decision}
\end{equation}
 
Under exact arithmetic, acceptance by the zero-tolerance certificate implies
\begin{equation}
B_0(x)=1 \;\Longrightarrow\;u_{\mathrm{cert}}\in\mathcal F_{\mathrm{exec}}(x)
\;\Longrightarrow\;\mathcal F_{\mathrm{exec}}(x)\neq\emptyset ,
\label{eq:exact_sufficiency}
\end{equation}
which follows directly from the witness acceptance test and is a sufficient, not a completeness, property; the implementation verifies the original hard set up to $\epsilon_{\mathrm{cert}}$. Except for an empty command box, $B_\epsilon(x)=0$ does not imply $\mathcal F_{\mathrm{exec}}(x)=\emptyset$. Unless otherwise stated, the experiments use the unscaled construction without a final command-box level, denoted Current; the box-final variant is examined in Sec.~\ref{subsec:sensitivity_results}.
 
\subsection{Certificate-Based Candidate Selection}
\label{subsec:certificate_based_candidate_selection}
 
Let $\mathcal C=\{c_1,\ldots,c_N\}$ be a finite candidate pool generated by an upstream planner, either a predefined motion library or an external geometric planner, and let $\mathcal C_G\subseteq\mathcal C$ be the subset accepted by its geometric validity criterion. For each $c\in\mathcal C_G$, a nominal kinematic rollout produces the checked states
\begin{equation}
    \hat q_{\kappa+1} =\hat q_\kappa+\Delta t_r\hat u_\kappa,\qquad \dot{\hat q}_{\kappa+1}=\hat u_\kappa,
    \label{eq:rollout}
\end{equation}
for $\kappa=0,\ldots,H-1$ with $\Delta t_r=0.05$~s and $H=8$, giving nine checked states over $0.40$~s. The provisional command $\hat u_\kappa$ follows nominal candidate tracking and is clipped to a propagation command box; it is not a propagated executor output, and the rollout integrates neither contact nor external-force dynamics. Certification bounds at each checked state use the executor interval $\Delta t=0.02$~s. Because the planner replans every $0.20$~s, we interpret prediction accuracy primarily over the first half of the horizon, where the median FR3 end-effector prediction error $0.20$~s ahead was 5.4~mm inside the interaction window. A candidate is admitted only if all of its checked states are certified:
\begin{equation}
    \mathcal C_{\mathrm{cert}}=\{c\in\mathcal C_G | B_\epsilon(\hat x_{c,\kappa})=1,\ \forall \kappa\in\{0,\ldots ,H\}\}.
    \label{eq:certificate_candidate_set}
\end{equation}
 
Both selectors use the same ranking cost $J$,
\begin{equation}
    c^*_G=\arg\min_{c\in\mathcal C_G}J(c),\quad c^*_\mathrm{cert}=\arg\min_{c\in\mathcal C_\mathrm{cert}}J(c),
    \label{eq:geometrically_valid_set}
\end{equation}
with $c^*_{\mathrm{cert}}$ defined only when $\mathcal C_{\mathrm{cert}}\neq\emptyset$; the two selectors differ only in their admissible sets. Let $\mathcal C_F\subseteq\mathcal C_G$ contain the candidates with $\mathcal F_{\mathrm{exec}}(\hat x_{c,\kappa})\neq\emptyset$ at all checked states, and let $\mathcal C_{\mathrm{cert},0}$ denote the set admitted by the zero-tolerance certificate. Then $\mathcal C_{\mathrm{cert},0}\subseteq\mathcal C_F\subseteq\mathcal C_G$, so $\mathcal C_{\mathrm{cert}}=\emptyset$ does not imply $\mathcal C_F=\emptyset$. Certification guarantees only local executor feasibility at the checked predicted states; it does not certify the rollout command itself, guarantee dynamic consistency across checked states, or guarantee successful closed-loop execution.

\section{EXPERIMENT}
\label{sec:experiment}
 
We evaluate the framework through numerical feasibility verification, controlled MuJoCo interaction experiments, and a planner-generated candidate-pool study using MoveIt~2 and OMPL.
 
\subsection{Numerical Verification Setup}
\label{subsec:numerical_verification_setup}
 
Numerical verification uses Franka Emika Panda and fixed-base RB-Y1 models in the Robotics Toolbox for Python (RTB)\cite{Corke2021RTB}; this Panda model is a separate instantiation from the MuJoCo FR3 model used in the interaction study, and the two sets of results are not interpreted as repeated evaluations of identical robot dynamics. For each robot, a deterministic grid combines 30 end-effector approach paths, 60 poses, three robot speeds $v\in\{0.1,0.3,0.6\}$, and three obstacle approach speeds $w\in\{0,0.25,0.5\}$, yielding 16,200 evaluations per robot. A configuration is geometry valid when $d_{\min}(q)\ge5$~mm and $q_{\min}\le q\le q_{\max}$. This deliberately partial criterion does not test self-collision, workspace bounds, velocity or acceleration limits, or command-box feasibility.
 
For a command-level reference label, we normalize the box as $u=\bar u+Dz$ with $D=\mathrm{diag}((h-\ell)/2)$ and $-1\le z\le1$, write $\tilde A=A_{\mathrm{hard}}D$ and $\tilde b=b_{\mathrm{hard}}-A_{\mathrm{hard}}\bar u$, and compute the signed normalized margin
\begin{equation}
   \begin{aligned}
         r^*=&\max_{z,r} &&r\\&\mathrm{s.t.}&&\tilde a_j^Tz\ge \tilde b_j+r\|\tilde a_j\|_2,\ \forall j,\\
         & &&-1+r\le z_i\le 1-r,\ \forall i,\\
         & && -1 \le z_i\le 1,\ \forall i
    \end{aligned}
\end{equation}
where $\tilde a_j^T$ is the $j$-th row of $\tilde A$. The margin bounds shrink the box for $r\ge0$, while $-1\le z_i\le1$ keeps the center inside the original box when $r<0$. No hard equality rows are active in this cohort. We classify $r^*\ge0$ as reference feasible and $r^*<0$ as reference infeasible; only nonnegative values are interpreted as a normalized inscribed-ball radius. For display only, Fig.~\ref{fig:validation}(b) reports $\min_{i:s_i>0}(1-|d_{\mathrm{cert},i}|/s_i)$; certification itself uses unscaled physical coordinates. The RTB certificate uses four construction levels: an empty initial level followed by workspace, self-collision, and obstacle rows.
 
\subsection{Interaction Simulation Setup}
\label{subsec:interaction_simulation_setup}
 
Controlled interaction experiments use MuJoCo with FR3 and fixed-base RB-Y1. Each run lasts 6.0~s with executor and planner periods of 0.02 and 0.20~s. During $t\in[2.5,3.1)$~s, a 10~N Cartesian force is applied along one of six world-frame directions. The physical force acts on the simulated dynamics, while the executor separately activates
\begin{equation}
    n^T_fJ_p(q)u\ge v_\mathrm{yield},
\end{equation}
where $n_f$ is the observed force direction and $J_p(q)$ is the translational Jacobian at the force-application point. $v_{\mathrm{yield}}$ is an application-level interaction-policy parameter rather than a universal safety threshold; we use $0.02$~m/s nominally and evaluate $0.005$--$0.05$~m/s. For predicted rollout states, FR3 inserts this row only after the observed force onset, whereas RB-Y1 uses the prescribed force schedule and therefore assumes known force timing.
 
For FR3, the force is applied at the center of mass of one of seven child links (42 conditions); the hard set contains the command bounds, workspace, obstacle, and interaction rows, and the soft hierarchy prioritizes end-effector tracking over posture. For RB-Y1, six locations over the proximal, intermediate, and distal regions of both arms yield 36 conditions; the hard set adds a support-polygon proxy and a cross-arm self-collision proxy, and the hierarchy prioritizes left, then right end-effector tracking, then posture. Each platform uses six candidates including the nominal motion and pose hold, with Cartesian offsets in $\pm X/\pm Z$ on FR3 and lateral $\pm Y$, a $+Z$ lift, and reduced forward progress on RB-Y1. FR3 uses five construction levels; RB-Y1 uses four or five depending on whether the interaction-obstacle row is active. Both selectors use the same platform-specific ranking cost at the terminal rollout state $H=8$,
\begin{equation}
J(c)=e_{\mathrm{EE}}(c)+10^{-3}\|\hat q_{c,H}-q_0\|_2,
\label{eq:ranking_cost}
\end{equation}
where $e_{\mathrm{EE}}$ is the terminal end-effector position error with respect to the nominal position at $t_{\mathrm{root}}+0.4$~s, taken as the RMS of the left- and right-arm errors for RB-Y1, and $q_0$ is the platform home configuration (7 joints for FR3; 20 for RB-Y1, comprising a 6-DoF torso and two 7-DoF arms); $J$ uses no certificate or reference-LP information.
 
Each condition uses 20 initial states generated by independent $0.01$-rad Gaussian joint perturbations. Seeds are shared across conditions and selectors and execution is deterministic thereafter, so intervals treat the 20 seeds as clusters rather than treating fixed-root events as independent. To isolate selection effects from closed-loop divergence, paired fixed-root evaluations at $t\in\{2.6,2.8,3.0\}$~s supply the geometry-based closed-loop state to both selectors with the same pool and ranking rule, yielding 2,520 FR3 and 2,160 RB-Y1 paired events plus 60 zero-force roots per platform; closed-loop executions are evaluated separately (1,720 FR3 and 1,480 RB-Y1 runs). When no candidate is certified, the executor continues to operate: FR3 holds a persistent end-effector pose anchor and RB-Y1 refreshes the current two-hand poses at each tick, both with zero feedforward. The command gap is then undefined, whereas nominal end-effector tracking error remains defined.
 
\subsection{Planner-Generated Candidate-Pool Experiment}
\label{subsec:planner_generated_candidate_pool_experiment}
 
Beyond the six-candidate libraries, we use 210 frozen FR3 roots from the geometry-based campaign at $t=2.8$~s (42 conditions, five seeds each). MoveIt~2 with OMPL RRTConnect serves only as a geometric candidate generator; it receives collision geometry and joint-position limits but no interaction constraint, certificate result, reference-LP label, or candidate score. Self-collision checking is disabled in the generated model, and retained paths satisfy joint-position limits, a workspace box, and sphere-proxy obstacle clearance. Accepted paths are resampled to nine waypoints by uniform RMS joint-space arc length, treated as duplicates within a $0.005$-rad RMS distance, and allowed up to 128 planning attempts per root to obtain 40 unique paths, yielding 8,400 candidates. For each root, 20 deterministic permutations define nested pools $\mathcal C_5\subset\mathcal C_{10}\subset\mathcal C_{20}\subset\mathcal C_{40}$. All retained paths use the same rollout and the FR3 ranking cost and are not executed in closed loop.
 
\subsection{Evaluation Metrics}
\label{subsec:evaluation_metrics}
 
At the paired fixed roots, selection outcomes are classified as unchanged, a transition to another moving candidate, a transition to pose hold, or no certified candidate. A transition is LP-qualified when the geometry-selected candidate has a reference-LP-infeasible checked state while the certificate-selected candidate exists and is LP feasible throughout. Planner--executor mismatch is measured by
\begin{equation}
    e_\mathrm{cmd}(t) =\|\dot q_\mathrm{plan,out}(t) -u_\mathrm{HQP}(t)\|_2,
\end{equation}
an interface-consistency diagnostic rather than a safety, task-performance, or overall control-quality metric. Because $e_{\mathrm{cmd}}$ is undefined when no candidate is selected, the primary comparison uses matched time support: timesteps for which the certificate run has a defined planner command are identified first, and the geometry-based gap is evaluated at exactly the same paired timesteps. We repeat the comparison after excluding pose-hold selections. Nominal end-effector tracking error is evaluated over all controller ticks and is not conditioned on command-gap availability.
 
For post-hoc exact-feasibility comparison, $\mathcal C_{\mathrm{LP}}\subseteq\mathcal C_G$ denotes the candidates that are reference-LP feasible at every checked state, with $c^*_{\mathrm{LP}}=\arg\min_{c\in\mathcal C_{\mathrm{LP}}}J(c)$; this baseline influences neither selector. When both selections exist, $\Delta J_{\mathrm{oracle}}=J(c^*_{\mathrm{cert}})-J(c^*_{\mathrm{LP}})$. For reached-state analysis we remove only the yielding row and compute $v_\mathrm{int}^{\max}(x)=\max_u n_f^T J_p(q)u$ subject to the remaining hard constraints and command bounds, giving the reserve $\rho_\mathrm{int}(x)=v_\mathrm{int}^{\max}(x)-v_\mathrm{yield}$, a local command-space diagnostic rather than a physical safety margin. Interaction confidence intervals bootstrap the 20 seed clusters, keeping all conditions and root times sharing a seed together; planner-pool intervals average the 20 permutations within each root and then bootstrap the 210 roots.
 
\subsection{Sensitivity and Ablation Setup}
\label{subsec:sensitivity_and_ablation_setup}
 
Offline evaluations on stored rollout states characterize sensitivity without rerunning the closed-loop campaigns. The interaction-policy parameter is varied over $v_{\mathrm{yield}}\in\{0.005,0.01,0.02,0.05\}$~m/s with candidate generation, ranking, rollout states, and all other constraints unchanged, and exact-LP feasibility and certificate admission are recomputed; no closed-loop claim is made for these settings. Level-order and box-final analyses also use frozen states, and the box-final construction is additionally evaluated in a separate RB-Y1 closed-loop ablation over the same 36 conditions and 20 seeds, with a frozen final-level box-face target of $10^{-10}$. Current is the production construction used for the complete cross-platform campaigns; box-final was introduced subsequently and is reported as a separate ablation. We report them as distinct operating points without interpreting either as universally superior.

\section{RESULTS}
 
\subsection{Geometric Validity and Executor Feasibility}
\label{subsec:geometric_validity_and_executor_feasibility}
 
\begin{table}[t]
\caption{Reference feasibility and certificate outcomes}
\label{tab:verification_results}
\centering
\footnotesize
\setlength{\tabcolsep}{3pt}
\renewcommand{\arraystretch}{1.5}
\begin{tabular}{@{}lccccc@{}}
\hline
Robot & Valid & Ref. feas.& Ref. infeas. & Certified & Uncert.\\
&&(\%)&(\%)&(\%)&(\%)\\
\hline
Panda   & 2,673 & 2,579 (96.48) & 94 (3.52)   & 2,469 (92.37) & 204 (7.63)\\
RB-Y1   & 2,412 & 1,711 (70.94) & 701 (29.06) & 1,517 (62.89) & 895 (37.11)\\
Overall & 5,085 & 4,290 (84.37) & 795 (15.63) & 3,986 (78.39) & 1,099 (21.61)\\
\hline
\end{tabular}
\vspace{-1.2em}
\end{table}
 
\begin{figure}[t]
    \centering
    \includegraphics[width=\columnwidth]{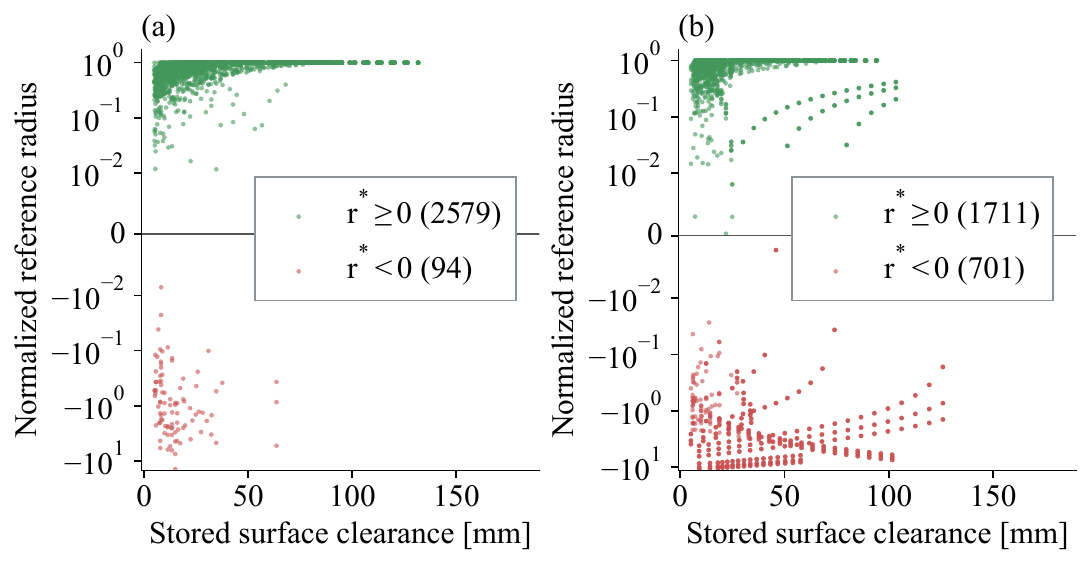}
    \caption{Stored surface clearance versus signed normalized reference margin $r^*$ for geometry-valid evaluations on Panda and fixed-base RB-Y1. Positive stored clearance can still yield $r^*<0$.}
    \label{fig:mismatch} \vspace{-1.2em}
\end{figure}
 
 The geometric-validity criterion retained 2,673 Panda and 2,412 fixed-base RB-Y1 cases out of 16,200 evaluations each (5,085 in total; Table~\ref{tab:verification_results}), of which 94 and 701 were reference   infeasible ($3.52\%$ and $29.06\%$). As Fig.~\ref{fig:mismatch} shows, positive stored surface clearance does not imply a nonnegative executor-feasibility margin.

 Among the 795, 764 (96.10\%) already violate a single hard row over the command box, that is, some row $j$ satisfies $\max_{\ell\le u\le h} a_j^T u<b_j$; only 31 (3.90\%) have every row individually satisfiable while their intersection is empty, with minimal infeasible subsets of size two in 29 cases and size three in two. This RTB cohort, however, contains no interaction row. In the MuJoCo interaction window, among the 2,520 FR3 fixed roots, a screen that tests rows individually passes 52 at which no executor-feasible command exists; all 52 arise from the combination of one obstacle-clearance row and the yielding row. Per-row screening would miss such coupled cases.
 
\subsection{Certificate Validation}
\begin{figure}[t]
    \centering
    \includegraphics[width=\columnwidth]{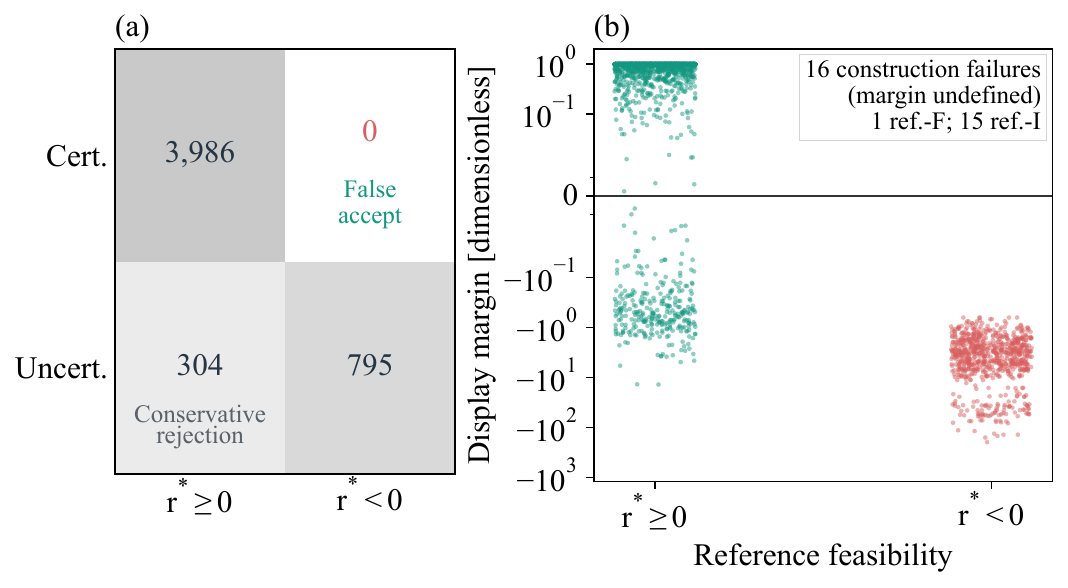}
    \caption{Validation of the feasibility certificate. (a) Cross-classification of certificate decisions and reference feasibility labels. (b) Certificate margin by reference-feasibility class, shown as the dimensionless display quantity of Sec.~\ref{subsec:numerical_verification_setup}. The 16 construction failures are annotated separately rather than plotted as finite margins.}
    \label{fig:validation}
\end{figure}
 
Among the 4,290 reference-feasible evaluations, 3,986 were certified and 304 remained uncertified, whereas none of the 795 reference-infeasible evaluations was certified (Fig.~\ref{fig:validation}). Of the 304, 303 produced a finite negative $m_{\mathrm{cert}}$ and one terminated during construction; of all 16 construction failures, 15 were reference infeasible, and no rejection arose from an empty command box or a post-margin residual failure. The incompleteness observed here therefore arises predominantly from constructed commands lying outside the physical command box rather than from failure of the level-wise construction. The absence of certified reference-infeasible evaluations is a finite-precision implementation check against the independent reference formulation, consistent with the sufficient direction of Eq.~\eqref{eq:exact_sufficiency} rather than a separate guarantee.
 
\subsection{Computational Cost}
\label{subsec:computational_cost}
\begin{figure}[t]
    \centering
    \includegraphics[width=\columnwidth]{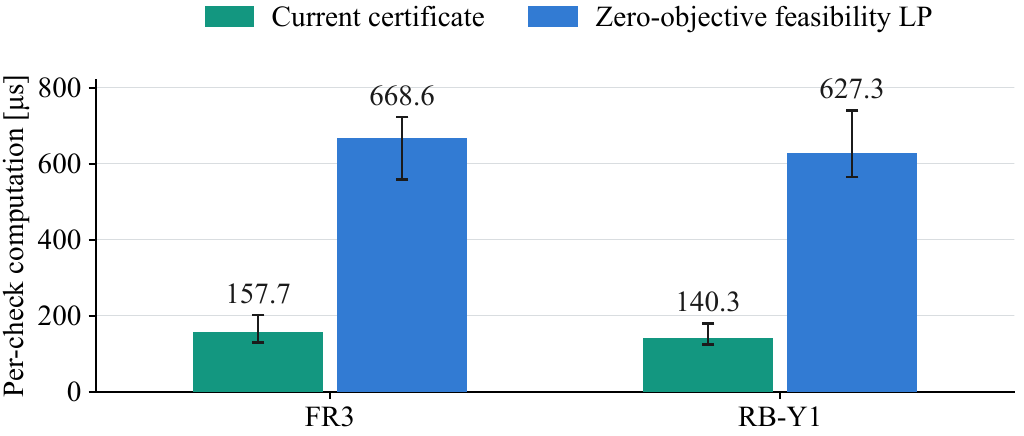}
    \caption{Median per-check computation time of the Current certificate and the zero-objective feasibility LP under the tested single-thread implementation and deployed state-level check scope.}
    \label{fig:computational_cost}
    \vspace{-0.6em}
\end{figure}
 
Under the tested single-thread implementation (Fig.~\ref{fig:computational_cost}), median per-check times over 480 preassembled command envelopes (240 per robot) were 158 and 140~$\mu$s for FR3 and RB-Y1, compared with 669 and 627~$\mu$s for a zero-objective feasibility LP over the same hard set solved with HiGHS via \texttt{scipy.optimize.linprog}. The certificate uses dense pseudoinverse operations on the same preassembled envelope; timing uses one CPU-pinned process with a single BLAS/OpenMP thread, three warm-up calls, and 21 interleaved timed calls per input, without LP warm starts. The zero-objective LP is distinct from the signed reference-margin problem defining $r^*$. These measurements characterize the tested implementations and solver configuration, not algorithmic complexity, and the two figures are not a controlled speed comparison: the certificate and the LP were exercised through different code paths without a common optimization effort.
 
\subsection{Interaction Results}
\label{subsec:candidate_selection_under_interaction}
 
\begin{table}[t]
\caption{Interaction-window closed-loop behavior\\(RB-Y1 EE: left / right)}
\label{tab:closed_loop_summary}
\centering
\footnotesize
\setlength{\tabcolsep}{2.8pt}
\renewcommand{\arraystretch}{1.05}
\begin{tabular}{@{}l l c c c c c@{}}
\hline
&Method &Coverage&\begin{tabular}{c}
     Matched\\gap\end{tabular}& \begin{tabular}{c}
        Tracking \\Error
     \end{tabular}&$\rho_\mathrm{int}^{5\%}$&\begin{tabular}{c}
     Neg.\\reserve
\end{tabular}\\ & &[\%]&[rad/s]&[mm]&[m/s]&[\%]\\
\hline
\multicolumn{7}{@{}l}{\textbf{FR3}}\\
& Geom. & 100.00 & 0.2551 & 14.767 & $-0.0200$ & 30.3 \\
& Current & 59.19  & 0.1920 & 19.477 & $-0.0200$ & 30.3 \\
\hline
\multicolumn{7}{@{}l}{\textbf{RB-Y1}}\\
& Geom. & 100.00 & 0.2191 & 5.360 / 6.468  & 0.0180 & 0.37 \\
& Current & 82.43  & 0.1831 & 13.082 / 13.365 & 0.0175 & 0.38 \\
\hline
\end{tabular}
\end{table}
 
\begin{figure}[t]
    \centering
    \includegraphics[width=\columnwidth]{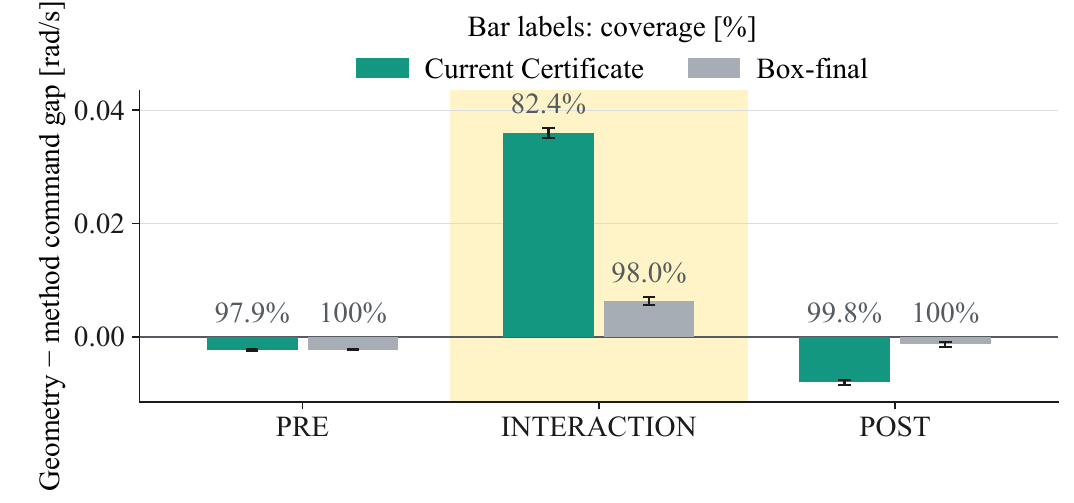}
    \caption{PRE / INTERACTION / POST decomposition of the Geometry-minus-method command gap for RB-Y1, the platform with box-final closed-loop data. Bar labels give planner-command coverage in each segment.}
    \label{fig:segment_gap}\vspace{-1.2em}
\end{figure}
 
Certificate-based admission changed the selected candidate in $35.99\%$ of FR3 and $45.92\%$ of RB-Y1 events, including transitions to pose hold in $11.15\%$ and $33.61\%$; LP-qualified transitions occurred in $35.16\%$ and $17.59\%$. The post-hoc exact-LP baseline clarifies the no-certified outcomes. For FR3, certified and LP-feasible availability were $54.29\%$ and $55.44\%$, and only 29 of the 1,152 no-certified events retained an LP-feasible alternative. For RB-Y1, availability was $85.09\%$ versus $100\%$, and all 322 no-certified events contained an LP-feasible candidate. At the nominal $v_{\mathrm{yield}}$, certificate incompleteness is therefore modest for the FR3 library but pronounced for RB-Y1. Conditional on a certified selection being available, it coincided with the minimum-cost LP-admitted selection in $88.38\%$ (FR3) and $60.55\%$ (RB-Y1) of events, with a median $\Delta J_{\mathrm{oracle}}$ of zero and means of $9.28\times10^{-4}$ and $6.77\times10^{-3}$. Conditioned on the events where the two selections differ, the mean rises to $8.0\times 10^{-3}$ and $1.7 \times 10^{-2}$, so the global median of zero reflects the frequency of agreement rather than the cost of disagreement.
  
Table~\ref{tab:closed_loop_summary} summarizes the interaction window. Coverage is the fraction of interaction-window controller ticks with a defined planner command, and command gaps use the matched support of Sec.~\ref{subsec:evaluation_metrics}, so the geometry rows are evaluated on the same restricted support, whereas nominal end-effector errors use all interaction-window ticks. The mean matched command gap was lower by $24.7\%$ and $16.4\%$ (paired differences 0.0631, seed-cluster $95\%$ CI $[0.0585, 0.0677]$; and 0.0360, $[0.0351, 0.0369]$), accompanied by reduced coverage and increased nominal end-effector error. These intervals reflect seed variation only. An implementation-level sensitivity of the RB-Y1 closed-loop difference, described among the limitations in Sec.~\ref{sec:conclusion}, is several times larger than these intervals in the six conditions where it was measured. The reached-state interaction-feasibility reserve showed no consistent advantage: the fifth-percentile reserve was identical on FR3 and differed by 0.52~mm/s on RB-Y1, and about $30\%$ of FR3 interaction-window states had negative reserve under both methods because the command box alone could not supply the prescribed yielding velocity.
 
The difference persists after excluding pose-hold selections. On moving-candidate timesteps the matched gaps were 0.2426 versus 0.1857 (FR3) and 0.1865 versus 0.1619 (RB-Y1), differences of $23.5\%$ and $13.2\%$, so pose hold does not fully account for the difference. The FR3 difference was heterogeneous: on moving-candidate timesteps the certificate gap was lower in 16 of 42 conditions, the median condition-level difference was near zero, and weighting conditions equally rather than timesteps reduces the FR3 moving-candidate difference from 0.0569 to 0.0316~rad/s. As shown in Fig.~\ref{fig:segment_gap}, the between-method difference is confined to the interaction window for RB-Y1: the Geometry-minus-Current gap is $-0.0024$ in PRE and $-0.0080$ in POST. Certificate-based admission is therefore associated with a lower command gap during the evaluated interaction window rather than with a general full-run reduction.

\subsection{Planner-Generated Candidate Pools}
 
Across the 210 frozen FR3 roots, certified availability increased from $36.6\%$ at $N=5$ to $52.9\%$ at $N=40$, tracking reference-LP availability ($38.6\%$ to $55.2\%$); at $N=40$, $2.4\%$ of pools had an LP-feasible alternative without a certified candidate, and LP-qualified transitions increased from $28.1\%$ to $44.3\%$. No certified candidate among the 8,400 was reference-LP infeasible. Under the tested implementation, filtering cost increased from 7.50 to 59.80~ms, compared with 34.15 to 273.28~ms for the LP.
 
\subsection{Sensitivity and Offline Ablations}
\label{subsec:sensitivity_results}
 
As $v_{\mathrm{yield}}$ increases from 0.005 to 0.05~m/s, FR3 certified availability decreases by 54.5 percentage points and reference-LP availability by 50.8 points, so certified availability as a fraction of reference-LP availability falls from $98.3\%$ to $78.8\%$. For RB-Y1 the corresponding decreases are 57.3 and 19.6 points, and the fraction falls from $100\%$ to $53.1\%$.
 
Appending the centered physical command-box faces as a final construction level recovered 244 of the 304 reference-feasible evaluations left uncertified by Current, losing no certified evaluation and accepting no reference-infeasible one. Of the remaining 60 ($1.40\%$), 59 failed at the final box level, where the activated projected system could not be satisfied within the construction tolerance, and one had already terminated at the obstacle level. Across four numerical-rank criteria their projected systems retained materially nonzero relative residuals (median 0.55), indicating projected-target inconsistency under the tested activation sequence rather than sensitivity to a numerical-rank cutoff. On the stored controlled states, box-final raised certification among reference-feasible states from $96.65\%$ to $99.43\%$ (FR3) and from $80.17\%$ to $95.07\%$ (RB-Y1) with zero false accepts, and alternative level orderings left FR3 certification unchanged and changed RB-Y1 certification by at most 0.056 percentage points.
 
In the separate RB-Y1 closed-loop ablation, box-final increased planner-command coverage from 82.43\% to 97.96\% and reduced mean left/right end-effector error from 13.08/13.36 to 8.42/8.99~mm. On the 17,625 interaction-window ticks where both variants have a defined planner command, however, Current has the lower command gap (0.1834 versus 0.2107~rad/s; seed-cluster 95\% CI for the paired difference $[-0.0284,-0.0264]$, reflecting seed variation only), also in the descriptive subset of 9,625 ticks where both select a moving candidate ($-0.0226$~rad/s) despite a mean planner-command norm about 20\% larger. The two constructions thus trade admission coverage against planner--executor interface consistency rather than one uniformly dominating the other.

\section{CONCLUSION}
\label{sec:conclusion}
Geometric validity does not imply that a downstream executor can issue any command at all: in 795 of 5,085 geometry-valid numerical evaluations the executor hard set was empty, mostly because of single hard rows that the geometric criterion does not check, and in the MuJoCo interaction window emptiness also arose from coupled rows that per-row screening would miss. The proposed certificate passes this information upstream without changing candidate generation, ranking, or the executor.

The interaction experiments clarify what this certificate does and does not provide. Admission changed the selected motion and was associated with a lower matched command gap during the forced-interaction window on both platforms, including on moving-candidate timesteps, while nominal tracking error increased and exact-LP admission exposed substantial conservatism on RB-Y1. The reached-state interaction reserve showed no consistent advantage, so the command-gap difference should not be interpreted as a safety improvement. The MoveIt/OMPL study extends the admission analysis to externally generated FR3 candidate pools; it does not evaluate closed-loop execution, and self-collision checking is disabled in the generated model. These results also localize the remaining incompleteness. Adding command-box faces as a final construction level recovered most of the missed reference-feasible states without false acceptance; nearly all remaining failures occurred at that final level, and their projected-system residuals stayed materially nonzero across four numerical-rank criteria.
 
Several limitations qualify these results. Seed-cluster confidence intervals quantify sensitivity to the randomized initial states, not heterogeneity across the fixed interaction conditions. The certificate is state-wise and does not establish forward invariance or viability; the braking terms in the command bounds address joint-limit approach only. Candidate checks use a kinematic rollout rather than propagated HQP, contact, or external-force dynamics. This rollout is least accurate where admission matters most: the median FR3 end-effector prediction error 0.20~s ahead was 5.4~mm inside the interaction window and 1.0~mm outside it. Interaction-row propagation is also platform specific: FR3 inserts the row after observed force onset, whereas RB-Y1 uses the prescribed force schedule for predicted states. Closed-loop differences on RB-Y1 also carry an implementation-level sensitivity. Advancing the executor's tick-time argument to the next representable double-precision value shifts the measured gap difference by 15--16\% of its value in the six conditions where this was measured, without changing its sign. The mechanism is executor-specific: trajectory divergence accumulates continuously and, after a median of 121 controller ticks, crosses a soft-QP status boundary at which the executed command switches discontinuously to a hard-feasible witness. FR3 has no such path and the corresponding relative shift is below $10^{-10}$. Validation is simulation-only and uses a narrow scripted-force model. $v_\mathrm{yield}$ is an application-level interaction-policy parameter, and the interaction row assumes that the force direction $n_f$ is fully observed. Finally, the experiments do not separate the contribution of executor-feasibility information from that of additional certificate conservatism to the observed command-gap difference.
 
Future work will therefore prioritize active-set revision for inconsistent projected targets, compare against exact and margin-matched admission policies to separate feasibility information from conservatism, and extend the state-wise interface toward trajectory-level feasibility.

\end{document}